\documentclass[conference]{IEEEtran}
\IEEEoverridecommandlockouts
\usepackage{cite}
\usepackage{amsmath,amssymb,amsfonts}
\usepackage{algorithmic}
\usepackage{graphicx}
\usepackage{textcomp}
\usepackage{xcolor}
\def\BibTeX{{\rm B\kern-.05em{\sc i\kern-.025em b}\kern-.08em
    T\kern-.1667em\lower.7ex\hbox{E}\kern-.125emX}}
\usepackage{verbatim}
\begin{document}

\title{Analysis of Recent Trends in Face Recognition Systems
}

\author{\IEEEauthorblockN{K. S. Krishnendu}
\IEEEauthorblockA{\textit{System Administrator} \\
\textit{catchmysupport.com}\\
Kerala, India \\
krishnendu@catchmysupport.com}
}

\maketitle

\begin{abstract}
With the tremendous advancements in face recognition technology, face modality has been widely recognized as a significant biometric identifier in establishing a person’s identity rather than any other biometric trait like fingerprints that require contact sensors. However, due to inter-class similarities and intra-class variations, face recognition systems generate false match and false non-match errors respectively. Recent research focuses on improving the robustness of extracted features and the pre-processing algorithms to enhance recognition accuracy. Since face recognition has been extensively used for several applications ranging from law enforcement to surveillance systems, the accuracy and performance of face recognition must be the finest. In this paper various face recognition systems are discussed and analysed like RPRV, LWKPCA, SVM Model, LTrP based SPM and a deep learning framework for recognising images from CCTV. All these face recognition methods, their implementations and performance evaluations are compared to derive the best outcome for future developmental works.  
\end{abstract}


\section{Introduction}
Automated face recognition has always been an area of focus since faces serve as an essential part of outer-world communication by revealing attributes like a person’s gender, ethnicity, age, and emotions. Rather than simply detecting the face from an image, establishing a person’s identity by processing and comparing image datasets has endless opportunities in law enforcement, surveillance systems, and video processing applications. However, variations between images of the same face due to changes in facial expression, aging, or illumination, and similarities between different facial images, as in the case of identical twins and genetically related persons, adversely affect the performance of face recognition technology.

\begin{figure}[htbp]
\centerline{\includegraphics[width=8cm]{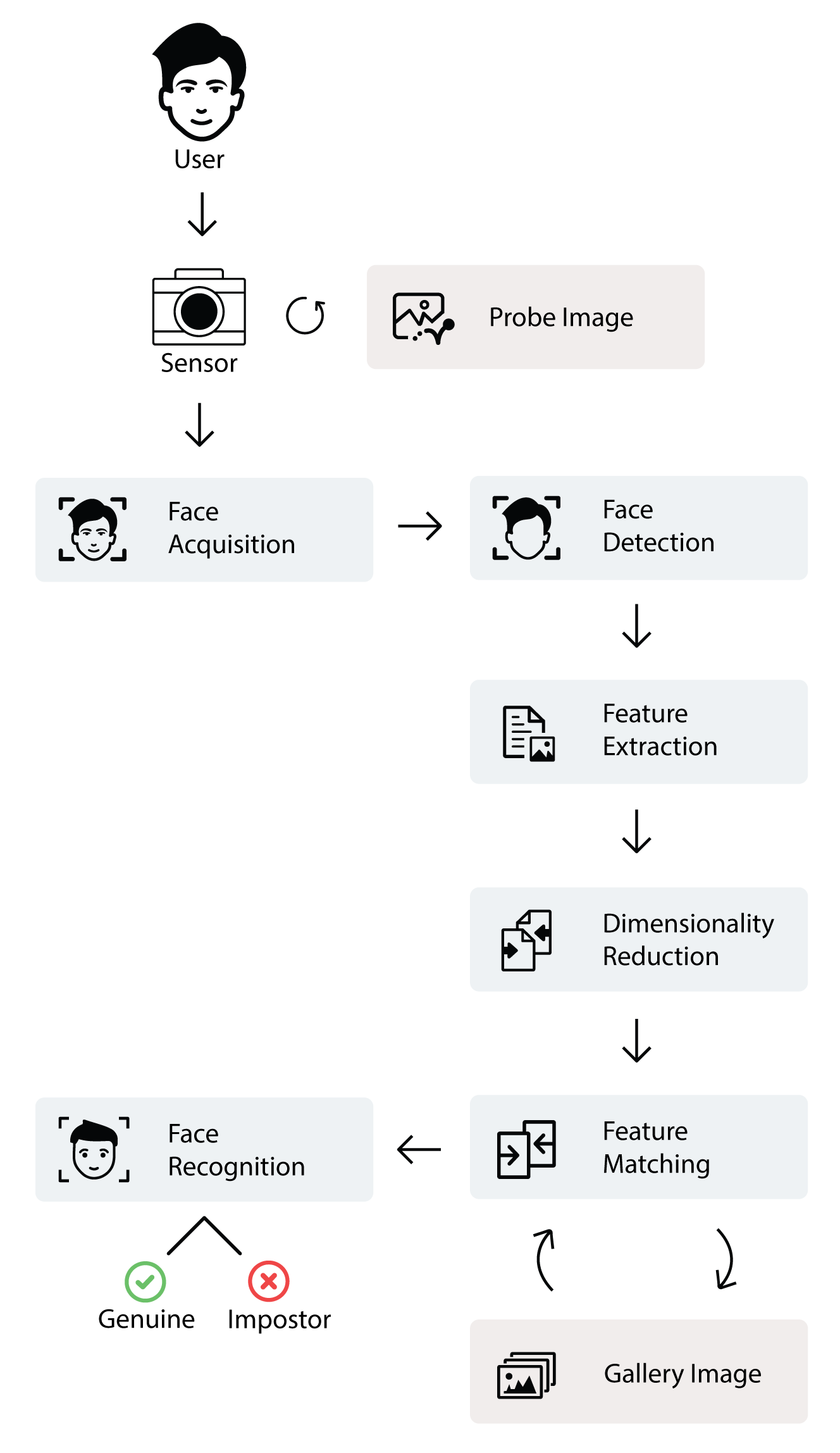}}
\caption{General Face Recognition Pipeline}
\label{fig1}
\end{figure}

Understanding the exact cognitive process involved in the recognition activity and training a machine for the same is an arduous task. So, each step involved in the design of a face recognition system is significant. A general face recognition model designed to recognize faces consists of image acquisition, enhancement, detection and extraction of features, dimensionality reduction and classification, and face recognition by matching probe image with the gallery sample(see ``Fig.~\ref{fig1}''). Image acquisition quality, face detection accuracy, and the robustness of feature extraction determine performance. Choosing the best image rendering technique can minimize any misalignment, rotational and pose changes, illumination changes, and blurriness. Appropriate face detection algorithms that efficiently determine the spatial extent and local features reduce false positive and false negative errors.  Also, among the three levels of facial features, extracting level 2 details from local regions using appropriate face descriptors is essential.

The scope of face recognition is based mainly on computer-vision-based and deep-learning-based approaches. Deep learning is an advanced area of research that requires high computational power and a huge dataset to get the desired result. Designing a deep learning model that extracts the feature set and classifies images is very complex. Whereas in the computer-vision-based approach, the preprocessing steps involved in face recognition, feature extraction, and classification are modified each time when the method involved is upgraded, or new technology is proposed. Discriminating analysis-based feature extraction efficiently reduces intra-class variations.

An analysis of the effect of race, gender, and skin tone of different demographic groups~\cite{kottakkal2021characterizing} the accuracy of face recognition systems suggests that there exists a disparity in accuracy even after training balanced datasets. The African American image cohorts have a higher false-match rate, and their Caucasian counterparts show a higher FNMR\cite{krishnapriya2020issues}. The impostor and genuine distributions are different between image cohorts. For a fixed decision threshold the recognition accuracy reduces due to higher false-match rates of African American and higher false non-match rates of Caucasian image cohorts\cite{S_2019_CVPR_Workshops}. When it comes to gender disparity, the recognition accuracy differs for men and women\cite{albiero2020analysis}. Female genuine distribution with lower similarity scores and female impostor distribution with higher similarity scores lower the accuracy of recognition\cite{albiero2020analysis}.Even though the female genuine distribution improves after removing facial occlusions like cosmetics, the imposter distribution recorded no change in the result. The brightness information available in the face skin region is correlative to the variation in matching accuracy\cite{albiero2020analysis}. For a mean range of skin brightness level, the image pairs show higher matching accuracy, and in all other cases, the result shows an increased false-match rate and increased false non-match rates\cite{wu2022face}.
\section{Related Work}
Detecting the face from an image and recognizing a person’s identity by comparing it with image datasets received much attention with the advancements in computer vision technology. Unlike any other biometric trait, training a machine to recognize human faces by learning the cognitive process involved in the recognition process has endless applications in law enforcement, intelligent video surveillance, and criminal identification and monitoring.

The developmental stages of face recognition technology can be classified into 3 phases:\cite{li2020review}
\begin{itemize}
    \item Early algorithms
    \item Artificial features and classifier stage
    \item Deep learning stage
    
\end{itemize}
In 1991, Turk and Pentland of MIT Media Laboratory discovered how to detect faces from an image. That led to the earliest instance of automated face recognition through a process called the principal component analysis(PCA)\cite{gottumukkal2004improved}. Like PCA, another linear feature extraction and dimensionality reduction method, LDA, was introduced, and that made use of label information to categorize different data as much as possible\cite{chintalapati2013automated}.

In 1995, Vapnik and Cortes proposed a Support vector machine (SVM), where an SVM classifier is trained to recognize the faces based on the features extracted from the embeddings of an input image\cite{cortes1995support}. Schapire proposed the Adaboost algorithm to improve the accuracy of any given learning algorithm by integrating different classifiers into a strong classifier based on some simple rules\cite{freund2003efficient}. Face recognition systems may fail to achieve the desired performance when the number of training samples in the image dataset is very small. In order to overcome these difficulties, Howland et al. proposed a method by performing linear discriminant analysis methods on small samples through a generalized singular value decomposition (GSVD)\cite{howland2006solving}.

Neural network-based algorithms are designed to simulate the process involved in face recognition by a human brain. A typical deep-learning approach is a branch of machine learning where the needed features are identified and classified automatically for the training process without involving any additional feature extraction step\cite{shalaby2021speaker}. A convolutional neural network (CNN) consists of neurons with shared weights and bias values. It uses the local data space and other features for optimizing the model structure by combining shared weights, local perception areas, and down-sampling of face images\cite{li2018implementation}.

\section{Methods Used in Recent Face Recognition Systems}
\subsection{Robust Prototype dictionary and Robust Variation dictionary(RPRV) method based Single Sample Per Person(SSPP) face recognition algorithm}
Currently, many face recognition systems use training sets with single sample per person (SSPP) images due to limited storage and privacy issues. But the recognition accuracy may be affected by strong intra-class variations between images of the same face, such as illumination changes, pose variations, aging, and occlusion. The sparse representation-based classification (SRC) technology, extended to generic learning method with the application of the Prototype and Variation dictionary model (P+V model), has been used for face recognition in recent years. However, this suffers due to within-class scatter problems in prototype dictionary and between-class scatter problems in variation dictionary. Upgrading the model with a robust prototype dictionary and robust variation dictionary construction (RPRV) method \cite{xue2022single} can solve the SSPP-related face recognition problems associated with intra-class and inter-class variations due to constrained and unconstrained environmental challenges. 

A robust prototype dictionary holding only the atoms with a bigger function value can improve the inter-class discriminative ability and there is no intra-class variation since gallery set has only a single sample per person.\newline
A robust variation dictionary with atoms having the smallest ratio of inter-class scatter to intra-class scatter reduce inter-class scatter and improve intra-class scatter in variation dictionary.\newline
The reconstructed sample by the robust prototype dictionary and the robust variation dictionary (RV) with smallest residual error gives the best recognition result.\newline RPRV model has following 3 steps:\newline
Dictionary learning:
\begin{itemize}
    \item Gallery images and generic images are combined to obtain a set of atoms to construct the robust prototype dictionary and robust variation dictionary (RPRV). 
    \item From a set of atoms obtained by dictionary learning, some effective atoms are selected with bigger inter-class scatter function values. 
    \item The reconstructed samples, regarded as atoms in the robust prototype dictionary and robust variation dictionary(RPRV) have the same label as the prototype used.
\end{itemize}
Dictionary construction:\newline
A robust prototype dictionary and robust variation dictionary are constructed with the following characteristics:\newline
Robust prototype dictionary:
\begin{itemize}
    \item Since the gallery set has only a single sample per person, there is no intra-class variation.
    \item But the inter-class scatter should be as large as possible to make the comparison between image samples easier.   
\end{itemize}
Robust variation dictionary:
\begin{itemize}
    \item Since the dictionary represents variation information of generic class samples, the intra-class scatter should be large and inter-class scatter should be as small as possible and which helps to improve the robustness of variation dictionary.  
\end{itemize}
If \emph{fg(j)} is inter-class variation of prototype dictionary and \emph{fv(j)} is inter-class variation of variation dictionary, then,

\emph{f(j)} =\emph{fv(j)} $-$ \emph{fg(j)}  where \emph{f(j)} will always be negative.

The smaller the \emph{f(j)} value, the better the prototype dictionary and the variation dictionary.
\newline
Face Recognition:
\begin{itemize}
    \item The probe sample, reconstructed by the robust prototype dictionary and the robust variation dictionary (RV) represents a reconstructed sample with the same label as the prototype used.
    \item The label with the smallest residual error between the reconstructed sample and the probe sample is the recognition result.
    
\end{itemize}

\subsection{Local Binary Pattern and Wavelet Kernel PCA(LWKPCA) method for face recognition}
Most of the hybrid face recognition systems use combined face features extracted by various feature extraction methods. Also, they implement the method of dimensionality reduction and a method of normalization on face datasets to minimize the variation effects to provide the best recognition results. But their performance degrades under adverse conditions. 

Local Binary Pattern and Wavelet Kernel PCA (LWKPCA)\cite{maafiri2022lwkpca} is a new feature extraction method to improve the robustness of FR systems. The method intends to improve recognition performance by extracting discriminant features with minimum recognition error. It involves a new feature grouping strategy based on Color LBP and Wavelet descriptor with a  non-linear subspace learning algorithm to extract discriminant features produced by the descriptor.

RKPCA:
\begin{itemize}
    \item It is a non-linear extension of RRQR PCA. In this method, the input image space is mapped into a feature space using non-linear transformation. Then using RRQR PCA factorization, the projection matrix is computed into that feature space.
\end{itemize}
CLWD(Color LBP and Wavelength Descriptor):
\begin{itemize}
    \item An input face image is represented as a discriminant vector structure by a novel feature grouping strategy by the Three-Level decomposition of Discrete Wavelet Transform (2D-DWT) and LocalBinary Pattern (LBP).
    \item For each color component R, G, and, B three-level decomposition of 2D-DWT is performed. 
    \item Then, Daubechies wavelet transformation is applied to Three-level decomposition from DB1 to DB4  to produce the four (LL3, LH3, HL3 and HH3) sub-images. But in the proposed method, they used only LL3 band to represent the facial image.
    \item The three sub-bands LL3(R), LL3(G) and LL3(B) are then converted into vectors structures such as LL3(R)v, LL3(G)v, LL3(B)v by a novel feature grouping strategy 
    \item Together with the histogram of LBP operator of the grayscale version of the input color image(Hr) are concatenated to generate the feature vector Fv of the proposed CLW Descriptor. Where r is the radius of the LBP operator.
    \item The feature vector Fv of each image is then used to generate feature training space for the face recognition model. 
    \item Even though there is no separate normalization module in the proposed method, features normalization is achieved by forming the matrix K by CLWD.
    \item The Lp-norm RBF kernel \[k_p(x_i,y_i)\]  produces a mapping to construct the feature space F, and, implicitly normalizes the two forms of features generated by two different feature extraction methods.
A non-linear dimensionality reduction is applied to the features generated by the proposed CLW Descriptor by using RKPCA algorithm.
Then, nearest neighbor classification is performed based on Euclidean distance to validate the robustness of the proposed method.

\end{itemize}
The LWKPCA method consists of two steps:
\begin{itemize}
\item Discriminating features extraction by the proposed Color LBP and Wavelet descriptor followed by the RKPCA algorithm to reduce the dimensionality.
\item A nearest neighbor classification algorithm based on the Euclidean distance is used for the classification of  feature vector of the input facial image with the feature vectors of the face dataset.
\end{itemize}
\subsection{Local-Tetra-Patterns based face recognition using Spatial Pyramid Matching}
A robust feature extraction method handles all intra-class variations, including pose variation, illumination changes, and covered faces, etc., without using any extra preprocessing techniques. 

Local-Tetra-Patterns-based face recognition using Spatial Pyramid Matching\cite{khayam2022local} involves 2 phases. Firstly a pattern map of the input image is created using LTrP, then the image partitioning is performed using SPM, then the histograms of the resultant image are created; afterward, max-pooling is executed on the obtained histograms to create the final feature vector, the last step is the classification based on obtained feature vector.

Feature Extraction and Formulation:
\begin{itemize}
    \item Pattern map of LTrP of the input image is calculated first in the feature extraction method. LTrP is mainly used for texture analysis as it can extract definite information from an image.
    \item SPM is utilized to partition each individual image into a number of blocks.
    \item Finally, the histogram for each block is calculated and the max-pooling method is applied to get final feature vector. Max pooling enhances discriminative power through dimensionality reduction of feature vectors and redundant value reduction.
\end{itemize}
Feature vector formation:
\begin{itemize}
    \item Concatenated histogram features carry much spatial information required for classification. So  by reducing redundant information, features become more discriminative, and the classification process will be enhanced due to the smaller feature vector.
\end{itemize}
Classification:
\begin{itemize}
    \item RKR based classifier is used for feature classification. RKR method helps to find maximum discrimination information stored in local features.
     \item Results obtained after compiling the percentage of the correctly identified test images represent the recognition accuracy.
     \end{itemize}

\subsection{Support Vector Machine (SVM) Model trained on Eigenvectors to perform face detection and recognition}

Monitoring and capturing students’ activities through online examination portals of respective educational institutions are significant in assessing the quality of teaching and learning. But variations such as noise, pose, and illumination changes affect the performance accuracy.
A reliable method like Eigenface, used to extract facial features through facial vectors trained by a Support Vector Machine algorithm, gives better recognition performance. 

The method\cite{geetha2021design} involves the following 3 steps:
\begin{itemize}
    \item At first, embeddings are extracted from the image dataset through facial vectors.
    \item The extracted embeddings are then trained by an SVM classifier. 
    \begin{itemize}
        \item It identifies a hyperplane in an N-dimensional space for classifying input data points.
       \item The data points closer to the hyperplane, called “Support Vectors,” are selected, which helps in maximizing the distance between the hyperplane and the data points with the help of a hinge function.
    \end{itemize}  
    \item Finally, the same SVM model is trained to recognize the face by comparing it with the data sets. The accuracy of face value is determined by a triplet loss function.
    
\end{itemize}
\subsection{A real-time framework for Automated Face Recognition in CCTV images}
The conventional CCTV system needs 24/7 manual checks, which is costly and time-consuming. Developing a machine learning and deep learning framework to automate face recognition in CCTV images with minimum manual interventions reduces computing time and improves performance. But the accuracy may be affected by rotation, scaling, and variations in light intensity. Images captured by a camera or video surveillance face several challenges due to noise, low resolution, and blurriness that degrades recognition performance. 

The framework\cite{ullah2022real} follows conventional steps of face recognition, including image acquisition, enhancement, face detection, feature extraction, and face recognition.
\begin{itemize}
    \item The facial image is captured using CCTV camera with appropriate specifications.
    \item Some pre-processing is performed to enhance the image using gray scale conversion and edge detection techniques. Edges of a filter are computed with the help of a hysteresis threshold.
    \item Viola–Jones algorithm is used to detect the face by distinguishing face and nonface regions. This state-of-art object detection algorithm records the highest face detection rate.
    \item PCA is used to extract features from an image and also helps in dimensionality reduction.
    \item Machine learning algorithms such as random forest, decision tree, K-nearest neighbor CNN are used to identify the best recognition result. And highest accuracy is obtained using CNN. 
\end{itemize}
\section{Discussions}
The performance analysis of the proposed RPRV method\cite{xue2022single} is done systematically, considering both constrained and unconstrained environments. The graphical analysis and tabular representation help the comparison better between RPRV method and other generic learning SSPP face recognition methods based on their accuracy percentage.
The proposed method\cite{xue2022single}can be further extended to use more deep learning features to improve efficiency. And the computational complexity of RPRV is smaller than the chosen four generic learning SSPP face recognition methods such as SRC, SSRC,S3RC and SGL. A brief theoretical background of the primary feature extraction methods needed while constructing the proposed method provided in this paper\cite{maafiri2022lwkpca} made it easier to get into the LWKPCA method.The proposed method doesn’t need a separate normalization module before dimensionality reduction like that in the state-of-art methods for feature extraction. Because the feature normalization is generated by CLWD by forming a matrix K. The Lp-norm RBF kernel \[k_p(x_i,y_i)\] implicitly normalizes the two forms of features generated by two different feature extraction methods. RKR based classification is much more efficient as they classify the non-linearly separable local features with maximum discrimination information. And SPM provides robust spacial discrimination information and max pooling reduces dimensionality and redundant information\cite{khayam2022local}. Since using a reliable method like Eigenface, used to extract facial features through facial vectors trained by a Support Vector Machine algorithm, the method\cite{geetha2021design} gives better recognition performance. The methods involved in the framework including Viola–Jones face detection algorithm, PCA-based feature extraction and dimensionality reduction, and CNN gives the best recognition result\cite{ullah2022real}.

The proposed method\cite{xue2022single} is not discussing the chances for the occurrence of inter-class similarities as in the case of identical twins and genetically related persons in prototype and variation dictionaries. Each column of the reconstructed probe sample matrix, merging gallery data matrix and variation matrix is high dimensional. So dimensionality reduction during dictionary construction will be a cumbersome task. Even though the paper\cite{maafiri2022lwkpca} mentions the nearest neighbor classification as a part of their proposed method to classify feature vector of input facial image by nearest neighbor classification algorithm using the Euclidean distance, a detailed explanation is not given.
The method\cite{khayam2022local} takes much processing time to complete due to the complexity of computing feature extraction and classification in two phases. The paper\cite{geetha2021design} didn't mention how recognition model handles varying illuminations and light intensities. CNN trained face recognition models that require high computational power are complex and the chances for error rate is also higher\cite{ullah2022real}.
\section{Conclusion}
Face Recognition is a vast area of research where a lot of efforts is needed to obtain the best possible outcome. In this paper, an analysis of recent recognition models and frameworks is made to formulate the efficient face detection and recognition algorithms and faster and reliable feature extraction methods for designing and developing a robust face recognition system. The performance evaluation of proposed feature extraction methods is verified with the help of the state-of-art image datasets. Some of the proposed models like LWKPCA are the upgradations of existing models. So it is clear that the face recognition technology is progressing at a much faster rate as researchers focus on introducing several new techniques to increase the recognition accuracy with low 
 processing cost and low computational time.
\bibliographystyle{ieeetr}
{\small
\bibliography{ref}}

% \bibitem{b1} G. Eason, B. Noble, and I. N. Sneddon, ``On certain integrals of Lipschitz-Hankel type involving products of Bessel functions,'' Phil. Trans. Roy. Soc. London, vol. A247, pp. 529--551, April 1955.
% \bibitem{b2} J. Clerk Maxwell, A Treatise on Electricity and Magnetism, 3rd ed., vol. 2. Oxford: Clarendon, 1892, pp.68--73.
% \bibitem{b3} I. S. Jacobs and C. P. Bean, ``Fine particles, thin films and exchange anisotropy,'' in Magnetism, vol. III, G. T. Rado and H. Suhl, Eds. New York: Academic, 1963, pp. 271--350.
% \bibitem{b4} K. Elissa, ``Title of paper if known,'' unpublished.
% \bibitem{b5} R. Nicole, ``Title of paper with only first word capitalized,'' J. Name Stand. Abbrev., in press.
% \bibitem{b6} Y. Yorozu, M. Hirano, K. Oka, and Y. Tagawa, ``Electron spectroscopy studies on magneto-optical media and plastic substrate interface,'' IEEE Transl. J. Magn. Japan, vol. 2, pp. 740--741, August 1987 [Digests 9th Annual Conf. Magnetics Japan, p. 301, 1982].
% \bibitem{b7} M. Young, The Technical Writer's Handbook. Mill Valley, CA: University Science, 1989.
% \end{thebibliography}
% \vspace{12pt}
% \color{red}
% IEEE conference templates contain guidance text for composing and formatting conference papers. Please ensure that all template text is removed from your conference paper prior to submission to the conference. Failure to remove the template text from your paper may result in your paper not being published.

\end{document}